\documentclass[letterpaper,twocolumn,fleqn]{article}

\usepackage{ist}
\usepackage{nameref} 
\newcommand{\mynamedref}[1]{'\emph{\nameref{#1}}'}
\usepackage{amsmath,amssymb}
\usepackage{url} 

\graphicspath{{./}} 
\DeclareGraphicsExtensions{.pdf,.png}

\title{Free-Space Detection with Self-Supervised and Online Trained Fully Convolutional Networks}

\author{ Willem P. Sanberg, Gijs Dubbleman and Peter H.N. de With \\Department of Electrical Engineering, Eindhoven University of Technology, the Netherlands}

\date{} 

\begin{document}

\maketitle

\thispagestyle{empty} 


\begin{abstract}
Recently, vision-based Advanced Driver Assist Systems have gained broad interest. In this work, we investigate free-space detection, for which we propose to employ a Fully Convolutional Network (FCN). We show that this FCN can be trained in a \emph{self-supervised} manner and achieve similar results compared to training on manually annotated data, thereby reducing the need for large manually annotated training sets. To this end, our self-supervised training relies on a stereo-vision disparity system, to automatically generate (weak) training labels for the color-based FCN. Additionally, our self-supervised training facilitates \emph{online} training of the FCN instead of offline. Consequently, given that the applied FCN is relatively small, the free-space analysis becomes highly adaptive to any traffic scene that the vehicle encounters. We have validated our algorithm using publicly available data and on a new challenging benchmark dataset that is released with this paper. Experiments show that the online training boosts performance with $5\%$ when compared to offline training, both for $F_\text{max}$ and $AP$.
\end{abstract}

\section{Introduction}
\label{sec:intro}

\begin{figure}[t]
\begin{center}
\includegraphics[width=\columnwidth]{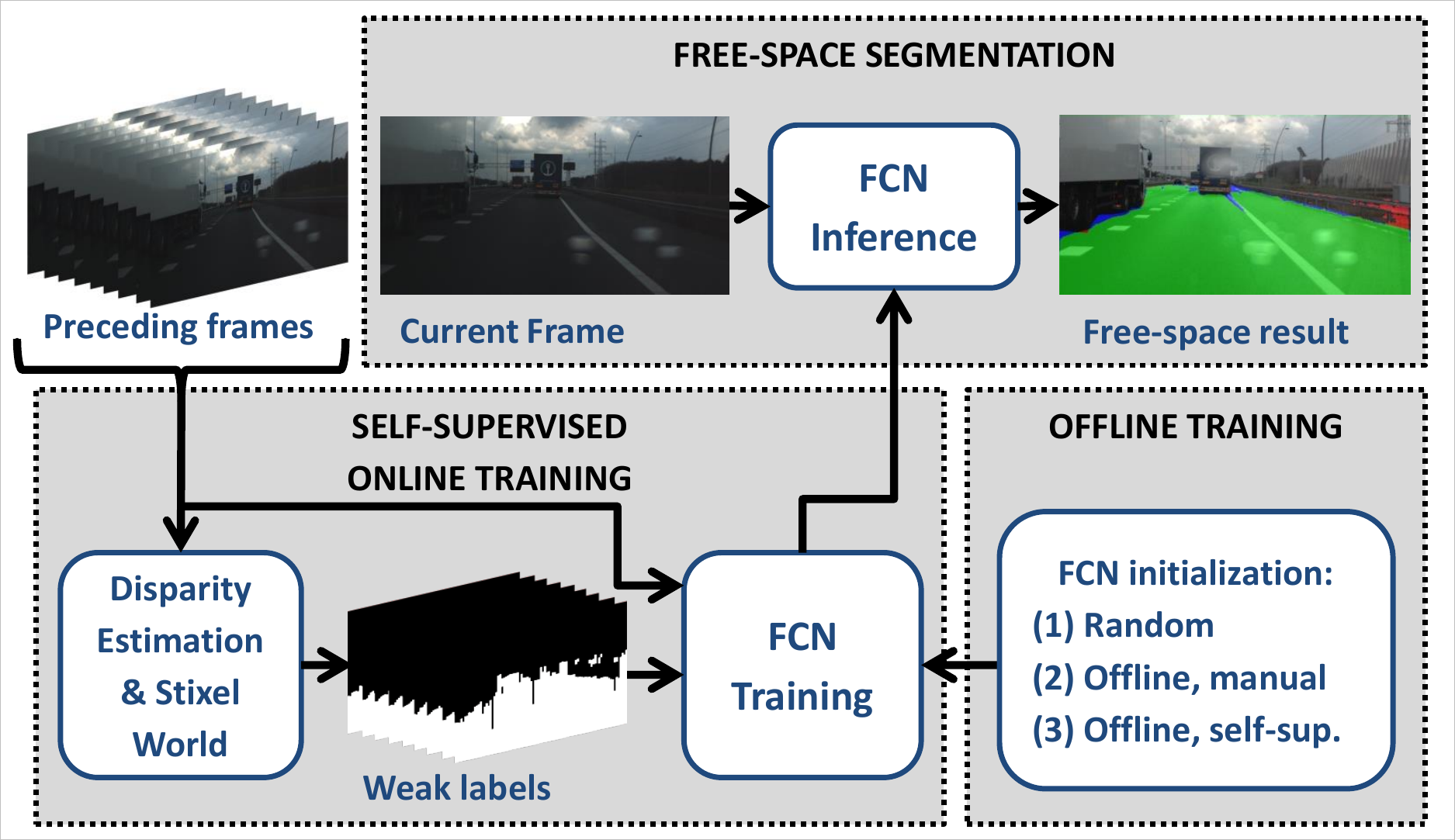}
\caption{Schematic overview of our free-space detection method with online self-supervised training. Detailed description in section~\mynamedref{sec:method}.}
\label{fig:scheme}
\end{center}
\end{figure}

In recent years, much research has been dedicated to developing vision-based Advanced Driver Assist Systems (ADAS). These systems help drivers in controlling their vehicle by, for instance, warning against lane departure, hazardous obstacles in the vehicle path or a too short distance to the preceding vehicle. As these systems evolve with more advanced technology and higher robustness, they are expected to increase traffic safety and comfort. A key component of ADAS is free-space detection, which provides information about the surrounding drivable space. In this work, we employ a Fully Convolutional Network (FCN) for this task and explore \emph{online} training in a \emph{self-supervised} fashion, to increase the robustness of the free-space detection system. Figure~\ref{fig:scheme} provides a schematic overview of our proposed framework, which will be described in detail in the~\mynamedref{sec:method} section.

Neural nets with deep learning are becoming increasingly successful and popular for image analysis. In the field of Intelligent Vehicles, many of the recent state-of-the-art algorithms rely on neural nets, mostly on Convolutional Neural Nets (CNNs). They excel in a wide variety of ADAS applications, such as stereo disparity estimation~\cite{zbontar2015stereo}, object detection for cars and pedestrians~\cite{RenHG015} and road estimation~\cite{mohan2014deep}\cite{Brust15VISAPP}.

In literature, training a neural net typically requires many data samples for proper convergence of the large amount of parameters and proper generalization of the classifier. Different strategies are adopted throughout the field to handle this. In image recognition and object detection problems in natural environments, a common method is to start with a net that is trained on a large and generic dataset, in either a supervised~\cite{Krizhevsky2012NIPS_4824} or unsupervised manner~\cite{hinton2006fast}\cite{erhan2010does}\cite{Dosovitskiy2014_NIPS_5548}. To apply it to a new task, one can remove the last layer of the net, which provides class confidences, and train a new one for the problem at hand. This exploits the observation that these pre-trained nets are a compact and yet rich representation of the images in general, since they are trained extensively on a broad visual dataset~\cite{zeiler2011adaptive}\cite{donahue2013decaf}\cite{zeiler2014visualizing}. An extension of this concept is not just retraining the last classification layer of a pre-trained net, but to also fine-tune a larger part or even the complete net with task-specific data~\cite{erhan2010does}\cite{Chatfield14}\cite{girshick2014rcnn}\cite{BransonVBP14}.

Scene labeling, in contrast to scene or object recognition, requires a per-pixel classification. Several strategies have been developed to go from global object detection to fine-grained segmentation, such as classification of sliding windows~\cite{sermanet2013overfeat} or image region proposals~\cite{girshick2014rcnn}, multi-scale CNNs combined with superpixels~\cite{Farabet2012PAMI}, or recurrent CNN architectures, which are a compact, efficient version of a multi-scale approach~\cite{pinheiro2014recurrent}. Recently, Fully Convolutional Networks (FCNs), have been employed for pixel-level segmentation~\cite{long_shelhamer_fcn}\cite{Brust15SUnW}. FCNs have several attractive properties in comparison to the aforementioned methods for scene parsing. For example, FCNs have no constraints on the size of their input data and execute inference in a single pass efficiently per image, instead of a single pass per superpixel, window or region~\cite{long_shelhamer_fcn}. Consequently, they do not require concepts like superpixel, region or multi-scale pre- or post-processing~\cite{long_shelhamer_fcn}. In the field of image segmentation, an additional interesting approach of training is weak supervision, where a limited set of related annotations are exploited for training. For example, the authors of~\cite{Pinheiro2015pixlev} train CNNs for pixel-level segmentation on image-level labels, since labels for the latter are more abundant than for the former. 

Even though weakly- or unsupervised training methods of CNNs are improving, they are currently still outperformed by fully supervised methods~\cite{Dosovitskiy2014_NIPS_5548}\cite{Pinheiro2015pixlev}. Together with the fact that creating large amounts of pixel-accurate training labels is inherently much work, we propose a middle-way in this paper: self-supervised training. If training labels can be generated automatically, the amount of supervised training data available becomes practically unlimited. However, this leads to a paradox, since it requires an algorithm that can generate the labeling, which is exactly the issue that needs to be solved. Therefore, we propose to rely on an algorithm based on traditional (non-deep learning) computer vision methods. This algorithm needs not to be perfect but at least sufficiently good to generate weak training labels. The goal is then that the FCN, trained with these weak labels, outperforms the traditional algorithm.

For next-generation ADAS, stereo cameras and multi-view cameras are an increasingly used sensor configuration. Stereo cameras provide insight into the geometry of the scene by means of the stereo disparity signal, which is valuable information for free-space detection. A state-of-the-art algorithm to distinguish free space and obstacles is the Disparity Stixel World~\cite{pfeiffer2012stixel}. It performs very well under favourable lighting conditions where the stereo estimation works reliably, but the algorithm is shown to have trouble under adverse conditions such as dim or very bright light~\cite{Pfeiffer2013}\cite{Sanberg2014ITSC}\cite{Sanberg2015ITSC}. We will use this algorithm to generate free-space masks and exploit these as weak training labels. We will rely on the generalization power of FCNs to deal with the errors in the weak labeling. In essence, we use a stixel-based disparity vision system to train a pixel-accurate free-space segmentation system, based on an FCN, and refer to this as self-supervised training.

As a further contribution, our proposed self-supervised training is enhanced by combining it with the aforementioned strategies of task-specific fine-tuning of neural nets. Since traffic scenes come in a wide variety (urban versus rural, highway versus city-center), with varying imaging conditions (good versus bad weather, day versus night), ADAS have to be both flexible and robust. A potential strategy is to train many different classifiers and to select the one that is most relevant at the moment (for instance, based on time and geographical location), or train a complex single classifier to handle all cases. In contrast, we show in this paper that it is feasible to fine-tune a relatively simple, single classifier in an online fashion. This is obtained by using the same self-supervised strategy as for offline learning, namely, based on generally correct segmentation by the disparity Stixel World. This results in automatically improved robustness of the free-space detection, as the algorithm is adapted while driving. Although our system does not yet operates in real-time, we deem this to be feasible in the near future, as the Stixel World system can execute in real-time and our FCN is relatively small, which allows for both fast training and fast inference.

Considering the overall approach, our work is also related to~\cite{Alvarez2012}, where automatically generated labels are exploited to train a CNN for road detection, which is applied as a sliding-window classifier. They also have an online component, which analyzes a small rectangular area at the bottom of the image (assumed road) and calculates a color transform to boost the uniformity of road appearance. The results of offline and online classifications are combined with Bayesian fusion. Our proposed work differs in several key points. Firstly, we do not need to assume that the bottom part of a image is road in the online training step, which is often an invalid assumption in stop-and-go traffic, since we exploit the stereo disparity as an additional signal. Secondly, their offline and online method is a hybrid combination of supervised and hand-crafted features, whereas our method can be trained and tuned in a fully end-to-end fashion, using a single FCN, while avoiding an additional fusion step. Thirdly, we do not require a sliding window in our inference step, since we use an FCN and not a CNN.

The remainder of this paper is structured as follows. Our self-supervised and online training strategies are described in more detail in section~\mynamedref{sec:method}. Our validation procedures are provided in section~\mynamedref{sec:experiments}, with a corresponding \mynamedref{sec:results} section. Finally, our research findings are briefly summarized in the \mynamedref{sec:conclusion} section.

\section{Method}
\label{sec:method}
In the following sections we will first explain the baseline FCN algorithm for image segmentation. After this, we will introduce our self-supervised and the corresponding online training strategies of the FCN in more detail.

\subsection{Fully Convolutional Network (FCN)}
\label{subsec:method:fcn}
The color-based segmentation algorithm used as a basis of our work is an FCN~\cite{long_shelhamer_fcn}. An FCN is a Convolutional Neural Network, where all fully connected layers are replaced by their equivalent convolutional counterparts. This adaptation transforms the net into a deep filter that preserves spatial information, since it only consists of filtering layers that are invariant to translation. Therefore, an FCN can process inputs of any size~\cite{long_shelhamer_fcn}. A challenge with this approach is that FCNs also typically contain several subsampling layers, so that the final output is smaller or of coarser resolution than the input image. To address this issue, the authors of~\cite{long_shelhamer_fcn} introduce skip-layers that exploit early processing layers, which have a higher resolution, to refine the coarse results in the final layer. In this way, the output resolution matches that of the input for pixel-level labeling.

For our experimentation, we have relied on the CN24 framework as described in~\cite{Brust15VISAPP}. Its extension to Fully Convolutional Networks is shortly introduced in~\cite{Brust15SUnW}. Provided that the context (road detection) and data (images captured from within a vehicle~\cite{Fritsch2013ITSC}) are comparable to our research, we adopt their network architecture and their recommendations about the optimal training strategy. The network consists of several convolutional, max pooling and non-linear layers: Conv ($7\times7\times12$); MaxP ($2\times2$); ReLU; Conv ($5\times5\times6$); ReLU; Full ($48\times$); ReLU; Full ($192\times$) + spatial prior; ReLU; Full ($1\times$) + tanh. The fully connected layers are interpreted and executed as convolutional layers by the CN24 library. The special feature of this network is the spatial prior, which is trained in the learning process as an integral part of the net, using the normalized positions of training patches. This spatial prior exploits the spatial bias that is certainly present in road or free-space segmentation in traffic scenes. We employ the recommended settings for training, based on image patches and without using dropout.

Note that our current work is not meant to offer an exhaustive test on optimizing the network architecture or the hyper parameters of the training process. We acknowledge the fact that our results may be improved by investigating that more properly, but the focus in this paper is to show the feasibility of self-supervised training and the additional benefits of our proposed online tuning in the context of free-space segmentation.

\subsection{Self-Supervised Training}
\label{subsec:method:self}

\begin{figure*}[tb]
\begin{center}
\includegraphics[width=\textwidth]{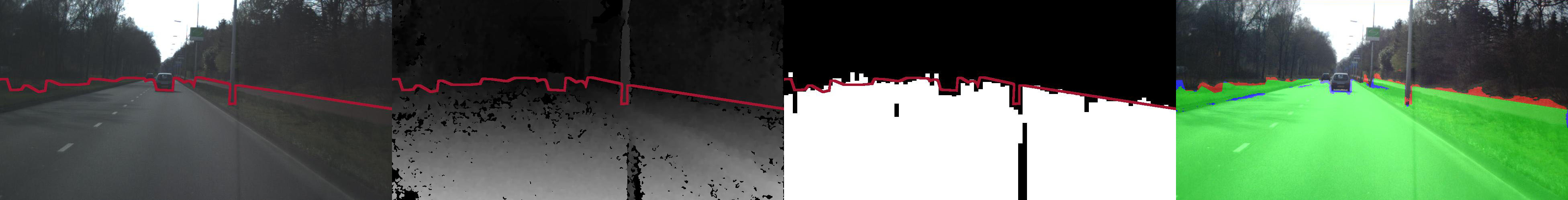}\vspace{1pt}
\includegraphics[width=\textwidth]{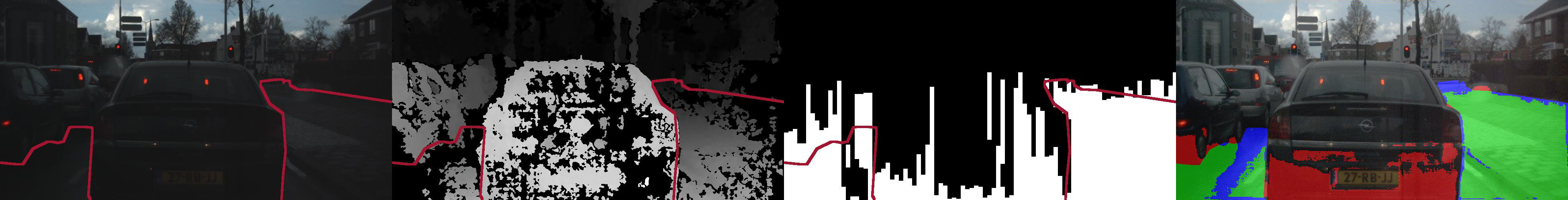}
\caption{Illustrating challenges in our self-supervised training: RGB images (taken under adverse but realistic circumstances), their disparity estimates and the stixel-based ground masks (all with ground-truth reference). Errors in the disparity lead to imperfect masks, so that these should be considered a \emph{weak} labeling. The preview of our results (last column) shows that we can deal with the errors to a large extent.}
\label{fig:weaklabeling}
\end{center}
\end{figure*}
Self-supervised training requires an algorithm that generates (weak) training labels. The training label generation algorithm is chosen to be an independent algorithm, which exploits an additional signal modality, namely stereo disparity. This disparity-based algorithm generates masks that indicate drivable surface. As these masks are estimates themselves and are not perfect, we say they represent \emph{weak} training labels. The reason to use a different modality (disparity) as basis for the weak training labeling than the modality (color) that is analyzed by the FCN, is to increase the chance that the trained algorithm can correct unavoidable errors in the weak labels, instead of stepping into the same pitfalls in difficult situations.

Stereo disparity is an attractive modality, since it is computationally inexpensive and yet provides relevant information in the context of free-space detection. We propose to analyze the disparity signal with the disparity Stixel World algorithm. This is a probabilistic framework that segments traffic scenes into vertically stacked, rectangular patches that are labeled as either ground or obstacle. The regularization within the Stixel World algorithm is mostly a-priori designed, exploiting the fact that disparity measurements facilitate metric/real-world reasoning, such as metric margins and gravity assumptions. By simplifying the representation of a scene into piecewise planar segments (flat for ground and fronto-parallel for obstacles), the segmentation can be formulated as a MAP estimation problem and can be solved efficiently using Dynamic Programming over columns with disparity measurements. The algorithm is highly parallel and can be executed in real-time~\cite{pfeiffer2012stixel}. In our system, we adopt the improvements to the disparity Stixel World algorithm mentioned in~\cite{Sanberg2014ITSC} (tuned transition probabilities, dynamic ground-plane expectation modeling and assuming a ground region below the image's field of view). The disparity-based ground and obstacle masks subsequently serve as the weak labels for the corresponding color image in our self-supervised training process.

Since this process can run automatically, we can generate (weak) labels for many frames. This facilitates training on images for which manual annotation is not available, so that we can enlarge the training set without much effort. Since many related deep-learning experiments have shown that training on more data can be beneficial (as discussed in the \mynamedref{sec:intro} section), we also experiment with self-supervised training on on an increased amount of frames. The challenge is that the generated weak labels will contain errors, as illustrated in Figure~\ref{fig:weaklabeling}, potentially hampering the training process. We rely on the generalization power of the FCN training process to deal with these inconsistencies in the labeling, which we can validate by comparing the results of our self-supervised training with the results of training on manually annotated frames.

\subsection{Online Training}
\label{subsec:method:online}

For online training, we adopt the training strategies as introduced in~\cite{Sanberg2015ITSC}. In that work, the stereo-disparity signal is analyzed for several frames, and the resulting segmentation labels are exploited to construct a color model of ground and obstacle regions. The color model is exploited to segment a new frame in the sequence with their color Stixel World algorithm. The color model they apply is a simple histogram over the predominant colors in an optimized, compact color space. In the process of generating the histogram, pixels are weighted with their real-world surface to balance image regions nearby and far away from the camera. With the histograms, class posteriors are generated using Bayes rule, which are subsequently applied in the MAP estimation of the color Stixel World~\cite{Sanberg2015ITSC}. Additional experiments over different color spaces showed that no single space is optimal for all frames~\cite{Sanberg2015PPNIV}. In other words, their color representation can be potentially improved by adapting it better to the imaging circumstances. Building further upon that observation, we propose to apply end-to-end learning in this work to exploit an FCN training algorithm for finding the representation of the image that is most relevant in the current situation.

A schematic overview of our experimental framework for free-space detection is shown in Figure~\ref{fig:scheme}. We train an FCN from scratch (with random initialization), or start with one of the offline trained models and tune the entire model with online data. By comparing these online strategies with results from solely offline training, we show the importance and added value of adapting the classifier online to the changing environment. If this adaptation can be realized in a reliable and realistic way, our free-space detection system improves without putting extra effort and computational power into training and executing a larger, more advanced FCN. By limiting the complexity of our system, real-time execution in a driving car becomes feasible in the near future.

\section{Experiments and Data}
\label{sec:experiments}
We discuss our experimental setup and data in the following subsections; firstly, the key experiment of comparing supervised and self-supervised learning of the FCN and secondly the comparison of online versus offline training. As a reference with general 3D modeling methods that do not rely on deep learning, we also compare against both the disparity and the color Stixel World methods. 

\subsection{Experiment 1: Supervised versus Self-Supervised Training}
\label{subsec:exp:selfsup}
To validate the feasibility of our self-supervised training method, we compare three FCNs that have an equal architecture but are trained with different data. The first model, $FCN_{\text{off,man}}$ is trained offline on manually annotated labels, as a reference result for offline, supervised training. The second model, $FCN_{\text{off,self}}$ is trained offline on the same frames as $FCN_{\text{off,man}}$, but now using automatically generated weak labels instead of the manual version. This model serves as a demonstration of offline, self-supervised training. Thirdly, we train a model in a self-supervised fashion on \emph{all} available frames in the dataset, including frames for which no manual labels are provided ($FCN_{\text{off,self-all}}$). This experiment tests the added value of training on additional data in our framework, which is realized efficiently because of the initial choice of fully self-supervised training.

\subsection{Experiment 2: Offline versus Online Training}
\label{subsec:exp:online}
We perform two key experiments to test the benefits of online training for our FCN-based free-space detection and compare this to the offline experiments of Section~\mynamedref{subsec:exp:selfsup}. Similar to Experiment~1, we train on different data while the architecture of our FCN is kept identical. Firstly, we train an FCN from scratch (with random initialization) on the weakly labeled preceding frames of each test frame, resulting in an $FCN_{\text{onl,scr}}$ for each test sequence. Additionally, we validate the benefits of online tuning. To this end, we initialize the net of each training sequence with one of the offline trained models (trained on either manual or self-supervised labels), resulting in an $FCN_{\text{onl,tun-man}}$ and $FCN_{\text{onl,tun-self}}$ for each test sequence. Note that the labels for the online training itself are always self-supervised, since the preceding frames of each sequence are not manually annotated.

Additionally, we perform an experiment to further analyze our online training method. Specifically, we show the power and benefit of '\emph{over-tuning}' for our framework. 
To this end, we test the online trained FCNs on test frames of different sequences than on which they were trained. By doing so, we can investigate the extent to which the online trained FCNs are tuned to their specific sequence. If the FCNs are over-tuned, we expect them to perform well if the training sequence and the test frame are aligned, but simultaneously expect them to perform poorly when they are misaligned. To validate this, we conduct three different misalignment experiments: (1) shift one training sequence ahead, (2) shift one training sequence back, and (3) randomly permutate all training sequences. 
Note that our data sequences are ordered in time, therefore, there can still be correlation between training sequences and test frames when shifting back or forth a single training sequence. We reduce this correlation as much as possible by randomly permutating all training sequences.

\subsection{Dataset}
\label{subsec:exp:data}
We utilize the publicly available data from Sanberg \emph{et al.}~\cite{Sanberg2014ITSC}\cite{Sanberg2015ITSC} as the training set for our offline training of the FCN. Combined, the data consists of 188 frames with manual annotations of free space. Additionally, the 10 preceding frames of each annotated frame are available, albeit without annotations. For our test set, we employ newly annotated data that is captured in a similar configuration and context as in~\cite{Sanberg2014ITSC}\cite{Sanberg2015ITSC}. The test set is released publicly with this work\footnote{available via \url{www.willemsanberg.net/datasets}} and consists of 265 hand-annotated frames of urban and highway traffic scenes, both under good and adverse imaging conditions. There is a large variety in scenes, covering crowded city centers, small streets, large road crossings, road-repair sites, parking lots, roundabouts, highway ramps/exits, and overpasses. To facilitate the online learning process, the 10 preceding frames of each annotated frame are provided as well (without manual labeling). The RGB frames are captured with a Bumblebee2 stereo camera from behind the windshield of a car. Both raw and rectified frames are available, as are our disparity images. These were estimated using OpenCV's Semi-Global Block-Matcher algorithm with the same settings as used in~\cite{Sanberg2015ITSC}. To the best of our knowledge at the time of writing, these publicly available and annotated datasets are unique in the aspects that they (1) readily provide preceding frames that are required to perform and evaluate online learning, and (2) consist of color stereo frames that facilitate our self-supervised training methods.

\section{Results}
\label{sec:results}
\begin{figure*}[tb]
\begin{center}
\includegraphics[width=\textwidth]{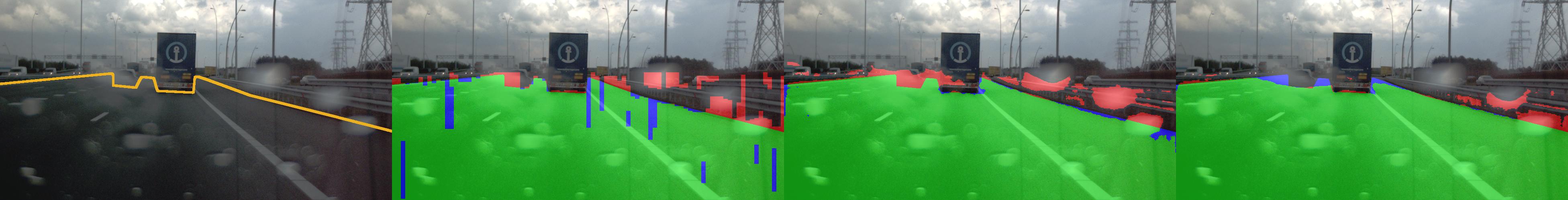}\vspace{1pt} 
\includegraphics[width=\textwidth]{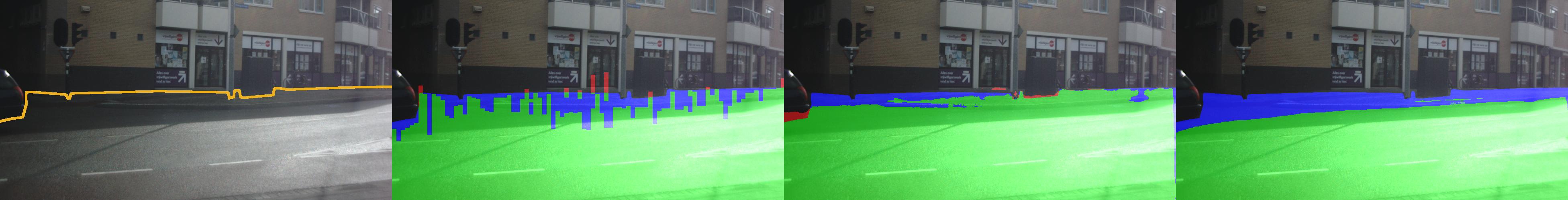}\vspace{1pt} 
\includegraphics[width=\textwidth]{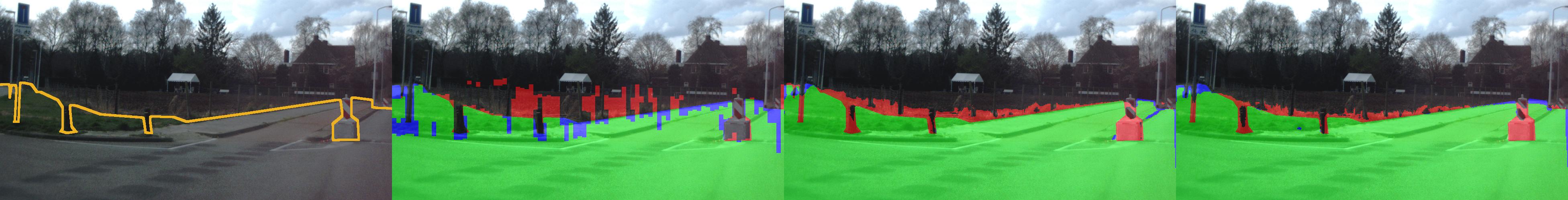}\vspace{1pt} 
\includegraphics[width=\textwidth]{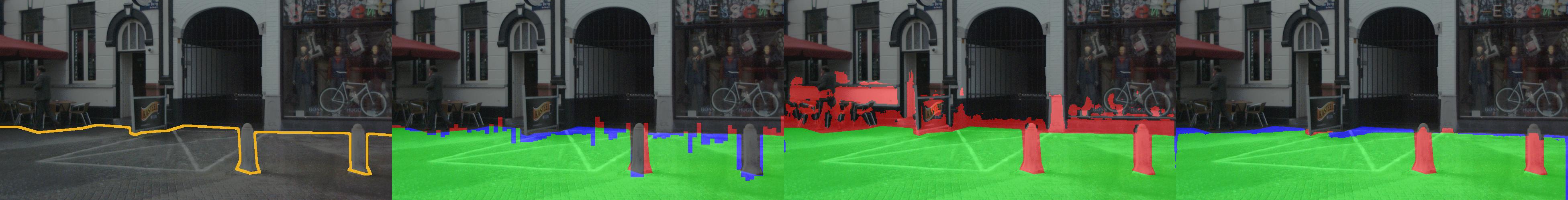}\vspace{1pt} 
\includegraphics[width=\textwidth]{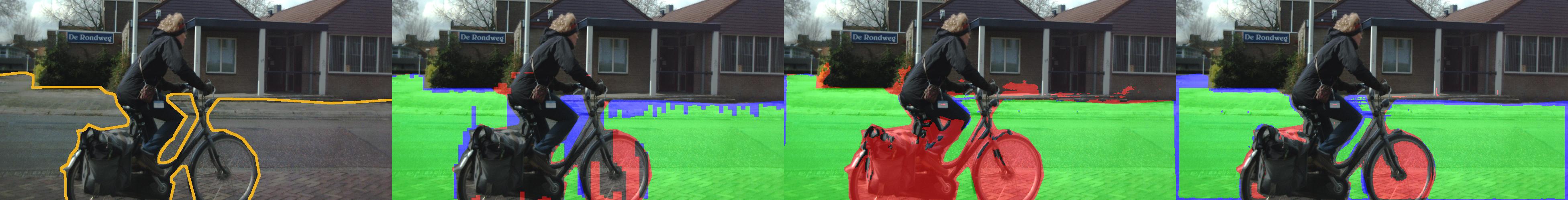}\vspace{1pt} 
\includegraphics[width=\textwidth]{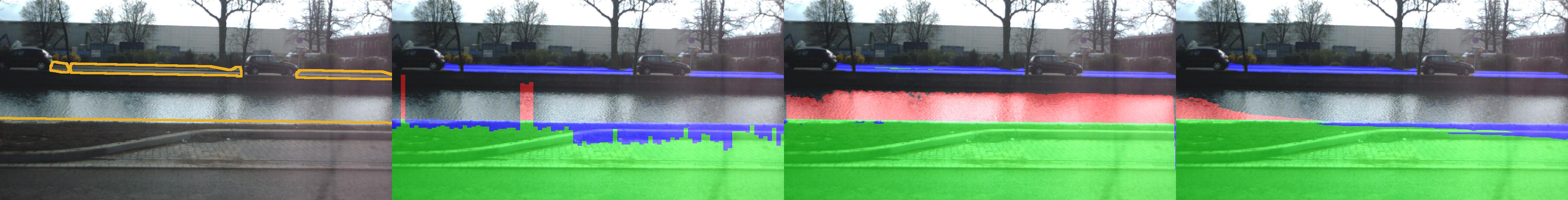}\vspace{1pt} 
\caption{Qualitative results on the frames shown in the leftmost column with hand-annotated freespace boundary. In the next columns, from left to right: Disparity Stixel World baseline; our FCN result with offline training on manual labels; our FCN result with online tuning. Colors indicate the freespace detections true positives (green), false negatives (blue) and false positives (red). Best viewed in color.}
\label{fig:Qual4H}
\end{center}
\end{figure*}

Figure~\ref{fig:Qual4H} shows qualitative results of our experiments. The first column contains the input color image (with free-space ground-truth annotation), and the second column the results of the disparity Stixel World baseline. The third and fourth column show results of our offline and online FCN-based methods, respectively. In the top three images, the offline-trained FCN detects less false obstacles than the Stixel World baseline. However, it performs worse in several important cases: it misses obstacles (such as the poles in the fourth row, the cyclist in the fifth row) and also classifies a canal as drivable (sixth row). In contrast, all-but-one images show that our online trained FCN outperforms both the Stixel World baseline and the offline training strategy. It segments the scene with raindrops on the car windscreen robustly, while the other results are also more accurate. The image in the fourth row shows an erroneous case: the online trained FCN does not detect the concrete poles, although they are present in the obstacle mask of the Stixel World algorithm.

For a more in-depth analysis, we also provide quantitative results. We adopt the pixel metrics as employed for the KITTI dataset~\cite{Fritsch2013ITSC}, which are shortly described here for completeness. We calculate Recall, Precision and the F-measure (the harmonic mean of Recall and Precision) in a Birds-Eye-View projection (BEV) of the image. Additionally, since our FCNs provide confidence maps, the maps need to be thresholded before they can be compared to the ground-truth annotation. To this end, we select the threshold that maximizes the F-measure, giving $F_{\text{max}}$. The metric $F_{\text{max}}$ is an indication of the optimal performance of the algorithm. To provide a balanced view, we also calculate the Average Precision $AP$, which captures the Precision score over the full range of Recall.

\begin{figure*}[tb]
\begin{center}
\includegraphics[width=0.9\textwidth]{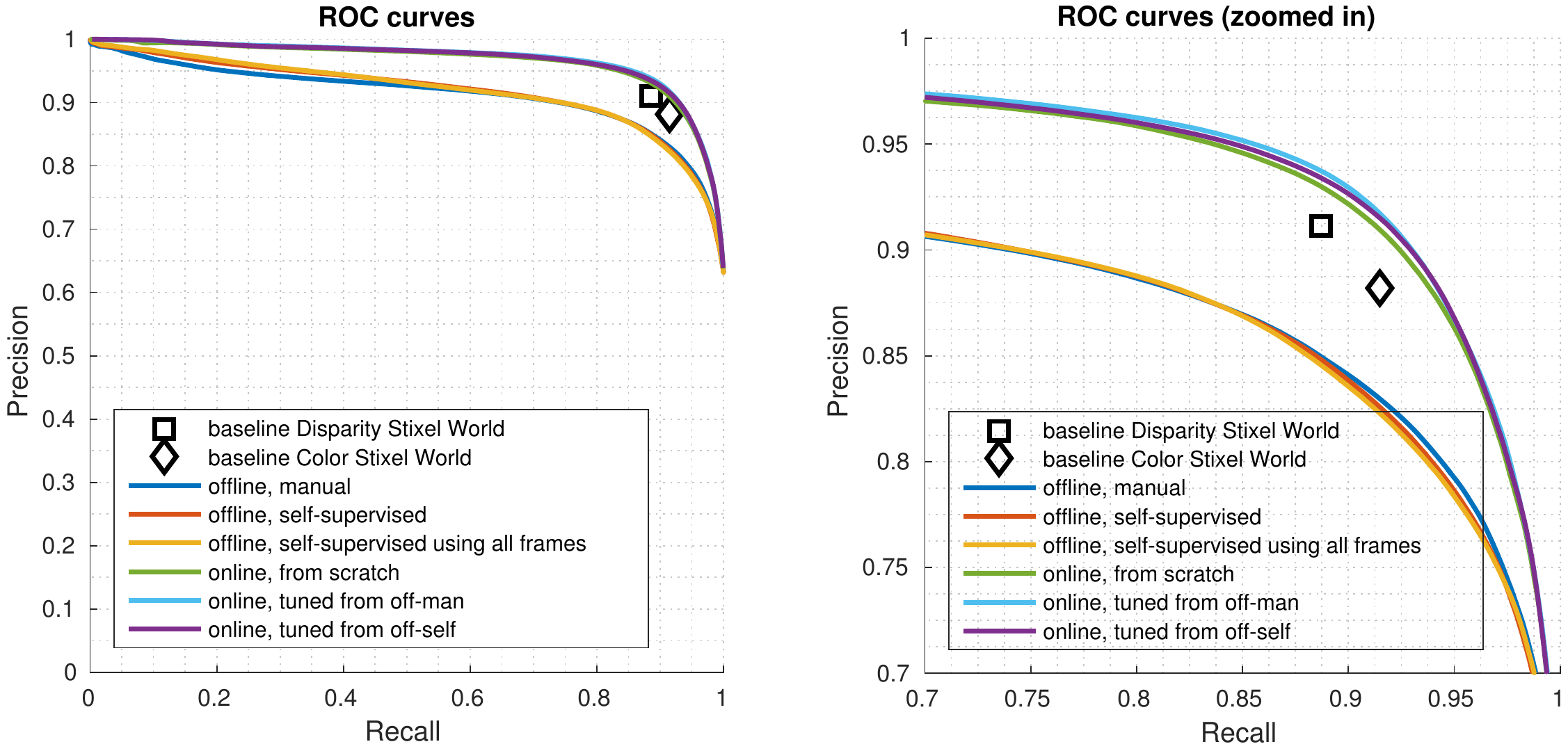}
\caption{ROC curves of our experiments after 10,000 training iterations. This figure is best viewed in color.}
\label{fig:ROC}
\end{center}
\end{figure*}

The Recall-Precision curves of our main experiments are shown in Figure~\ref{fig:ROC}, where the left graph shows the full range and the right graph provides an enlarged view of the top-right region (closest to the optimal point). For offline learning, the results of supervised (manual labels) and self-supervised (disparity-based labels) are nearly identical. This confirms the feasibility of self-supervised learning, as relying on weak labels does not hamper the performance of our system. Self-supervised training on more data did not lead to a clear improvement of the results in our experiments, as illustrated by the graph. This may show that our network is too small to exploit the additional data, or that the correlation within the new samples is too high to be informative. Considering our online training strategies, Figure~\ref{fig:ROC} clearly shows that these outperform the offline training over the full range of Recall, thereby confirming the qualitative results of Figure~\ref{fig:Qual4H}.

\begin{figure*}[tb]
\begin{center}
\includegraphics[width=0.9\textwidth]{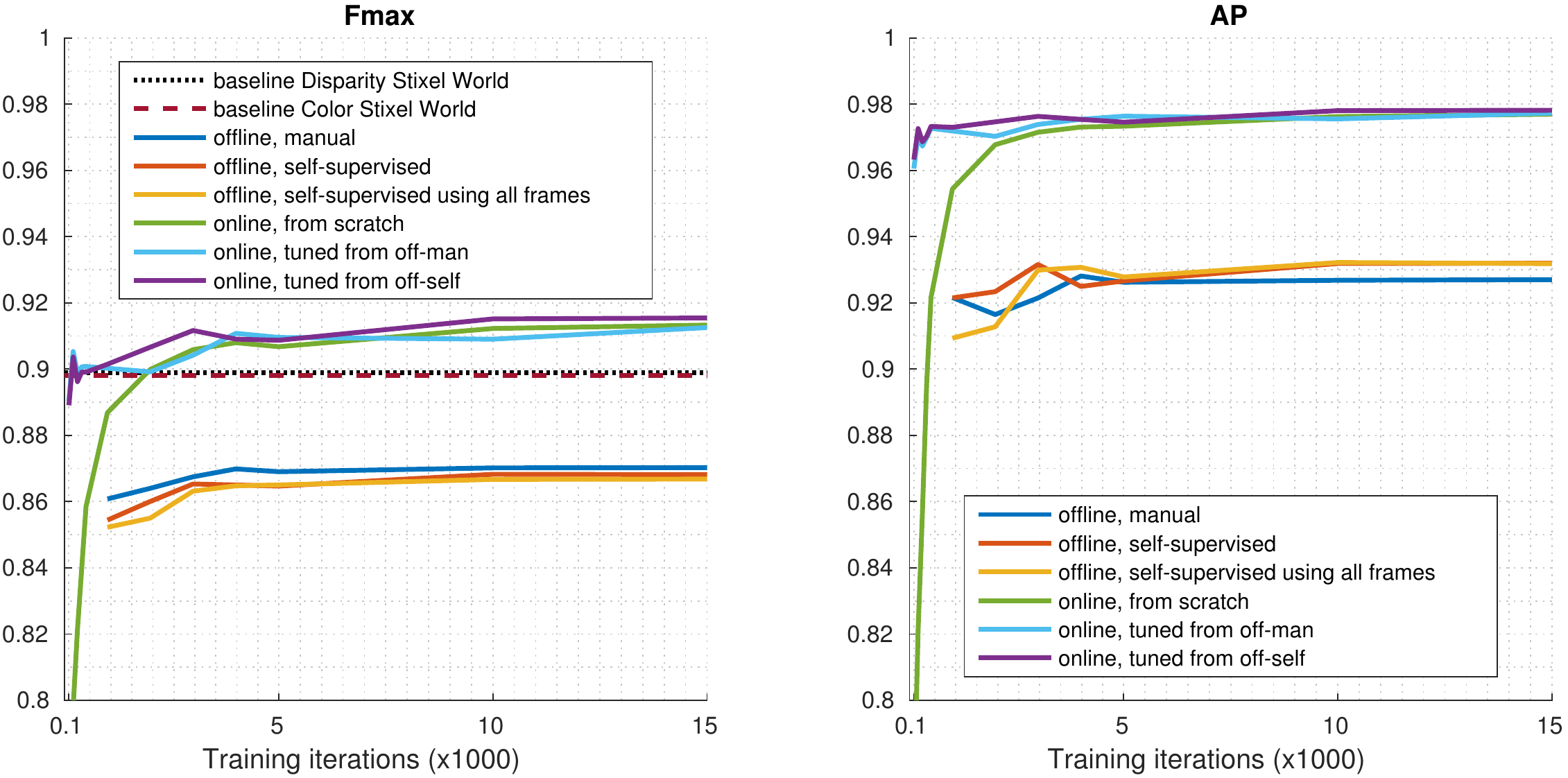}
\caption{Fmax and AP convergence over FCN training iterations. This figure is best viewed in color.}
\label{fig:QoverEp}
\end{center}
\end{figure*}

\begin{figure*}[tb]
\begin{center}
\includegraphics[width=\textwidth]{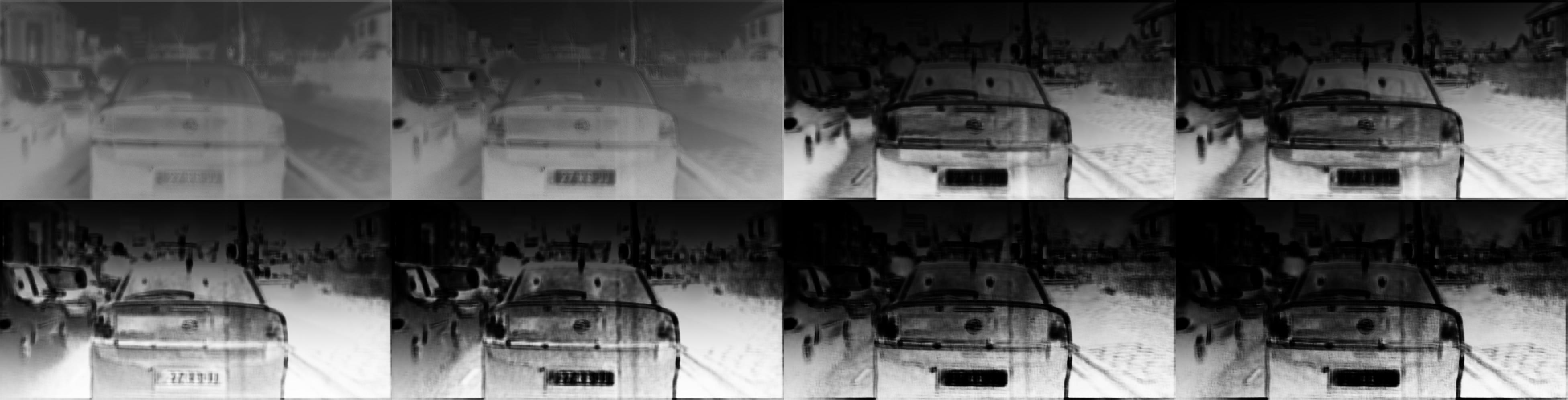}
\caption{Illustrations of convergence speed for online FCN training methods. Top row: $FCN_{\text{onl,scr}}$; bottom row: $FCN_{\text{onl,tun-self}}$; both after 100, 500, 5,000 and 10,000 iterations. Although both nets end up with similar masks, initializing the online FCN with a pre-trained net clearly speeds up the convergence. The input image is shown in the second row of Fig.~\ref{fig:weaklabeling}.}
\label{fig:confmasks}
\end{center}
\end{figure*}

\begin{figure*}[tb]
\begin{center}
\includegraphics[width=0.49\textwidth]{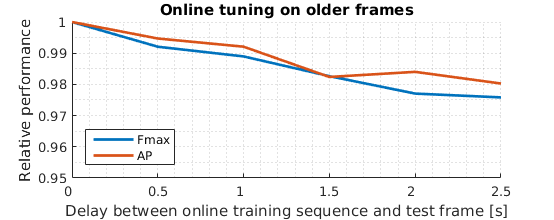}
\includegraphics[width=0.49\textwidth]{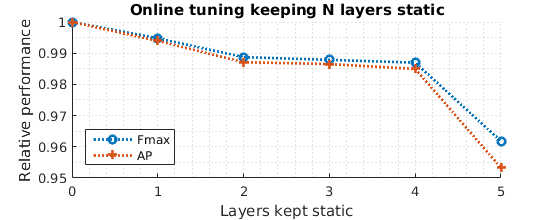}
\caption{Two different experiments on robustness of our online tuning strategy. Left: tuning on older frames. Right: tuning only the last couple of layers, with the first N static (N=0: tune all layers, N=5: no tuning; same as offline training only). }
\label{fig:scoreDrops}
\end{center}
\end{figure*}

The trends of our quantitative results over the number of training iterations are shown in Figure~\ref{fig:QoverEp}. The training converges after 5,000 to 10,000 iterations and the visible trends are consistent with the ROC curves in Figure~\ref{fig:ROC}: offline training is outperformed by online training and online tuning performs slightly but consistently better than online training from scratch. An important conclusion of the experiments is that the contribution of online-tuned training is most significant in the speed of convergence, and less relevant for the final result after convergence. More specifically, the models for which we apply tuning outperform offline methods and the baseline already after 100 iterations of training (which takes less than half a second on a GeForce GTX970 graphics card), whereas models trained from scratch need at least 500 iterations to match the offline FCN and more than 2000 to exceed the Stixel World algorithm. Figure~\ref{fig:confmasks} illustrates how the confidence masks of these two methods converge.

Additionally, we have conducted two different experiments on the robustness of our online tuning strategy. Firstly, we have tested the drop in performance as a function of the delay between the frames on which the online tuning is performed and the frame under analysis. The left graph in Figure~\ref{fig:scoreDrops} shows that the score drops only about $2\%$ with a delay of 2.5 seconds. Secondly, we validated the influence of the number of FCN layers that are tuned online on the free-space detection result. Our net has 5 layers with parameters, so we compared tuning all layers (regular online tuning) to tuning only the last 4, 3, 2 or 1 layers. Keeping all layers static is the offline training reference. The right graph in  Figure~\ref{fig:scoreDrops} shows that tuning only the final layer provides results within $1.5\%$ of the full tuning approach. Both experiments provide tradeoffs between tuning time and performance quality. 

The results of the misalignment of the training sequences and the test frames with the online-trained FCNs are provided in Table~\ref{table:quantshuff}. It is clear that the misalignment has a negative impact on the performance of the online training approach, as was to be expected. The scores drop even below that of the models that are trained offline, also for the FCNs that were initialized with offline pre-trained nets. As the online FCNs outperform all other methods when their training sequence and test frame are aligned, this validates our claim that the online training is giving the system flexibility to adapt to new circumstances, and that over-tuning can be exploited beneficially in the context of free-space detection for ADAS.

\setlength{\tabcolsep}{12pt}
\begin{table*}
\caption{Table 1. Results of online-training FCNs on different training sequences: aligned (normal), one sequence back or ahead in the (ordered) dataset (+1/-1), or permutated randomly. The drop in performance illustrates the adaptive power of the online tuning.}
\label{table:quantshuff}
\begin{center}
\renewcommand{\arraystretch}{1.4}
\begin{tabular}{|l||c||c|c|c||c|c|c|}
\hline
                 & offline & \multicolumn{3}{c||}{trained online from scratch}& \multicolumn{3}{c|}{tuned online from off-self}  \\
                 \cline{3-8}
                 & (manual)  & normal & +1/-1 & random                          & normal & +1/-1 & random   \\
\hline
$F_{\text{max}}$ & 0.87    & 0.91 & 0.83 & 0.79                               & 0.92 & 0.83 & 0.80 \\
$AP$             & 0.93    & 0.98 & 0.91 & 0.84                               & 0.98 & 0.92 & 0.86 \\
\hline
\end{tabular}
\end{center}
\end{table*}

\section{Conclusions}
\label{sec:conclusion}
We have shown that Fully Convolutional Networks can be trained end-to-end in a self-supervised fashion in the context of free-space segmentation for ADAS. The segmentation results are similar to a conventional supervised strategy that relies on manually annotated training samples. We expect that this result can be generalized to different end-to-end training algorithms, reducing the need for large amounts of manually labeled data. Furthermore, we have extended this result to show that it facilitates \emph{online} training of a segmentation algorithm. Consequently, the free-space analysis is highly adaptive to any traffic scene that the vehicle encounters. Experiments show that the online training boosts performance with $5\%$ when compared to offline training, both for $F_\text{max}$ and $AP$. In conclusion, we exploit the fact that our adaptive strategy is not required to generalize to a large amount of traffic scenes with a single detector. Hence, the detector can -and should- be 'over-tuned' on currently relevant data. In turn, this allows for a small FCN whose training converges fast enough to make real-time deployment feasible in the near future.



\section{Acknowledgments}
We happily acknowledge the assistance of C.-A. Brust with employing the CN24 library.




\vfill\pagebreak


\begin{biography}


Willem Sanberg received both his BSc. (2011) and his MSc.-degree (2013) in Electrical Engineering and a certificate in Technical Management (2011) from Eindhoven University of Technology in the Netherlands. His PhD research is aimed at improving the semantic understanding of 3D modeled environments for intelligent vehicles by developing efficient yet robust methods that can adapt themselves to the dynamic environment of everyday traffic. He works in (inter)national projects with both industrial and research partners.

Dr. Gijs Dubbelman is an assistant professor with the Eindhoven University of Technology and focuses on signal processing technologies that allow mobile sensor platforms to perceive the world around them. He obtained his PhD. research in 2011 on the topic of Visual SLAM. In 2011 and 2012 he was a member of the Field Robotics Center of Carnegie Mellon's Robotics Institute, where he performed research on 3-D computer vision systems for autonomous robots and vehicles.

Peter H. N. de With (MSc. EE) received his PhD degree from University of Technology Delft, The Netherlands. After positions at Philips Research, University Mannheim, LogicaCMG and CycloMedia, he became full professor at Eindhoven University of Technology. He is an (inter-)national expert in surveillance for safety/security and was involved in multiple EU projects on video surveillance analysis with the Harbor of Rotterdam, Dutch Defense, Bosch Security, TKH-Security, ViNotion, etc. He is board member of DITSS and R\&D advisor to multiple companies. He is IEEE Fellow, has (co-)authored over 300 papers on video analysis, systems and architectures, with multiple awards of the IEEE, VCIP, and EURASIP. 
\end{biography}

\end{document}